\documentclass{article}
\usepackage[utf8]{inputenc}
\usepackage{amsmath}
\usepackage{siunitx}
\usepackage{hyperref}
\usepackage{graphicx}
\usepackage[section]{placeins}

\begin{document}

\title{Review Paper: Inertial Measurement}

\author{William T. Conlin (wtconlin@gmail.com)}
\date{May 2017}

\maketitle

\begin{abstract}
Applications of inertial measurement units are extremely diverse, and are expected to see a further increase in number due to current trends in robotics as well as recent advances in Micro Electromechanical sensors (MEMS). The traditional method of inertial measurement has depended on costly, power-intensive, error-prone Inertial Measurement Units (IMUs) that represent a single point of failure. Promising areas of current research include methods for combining multiple redundant sensors, which collectively provide more accurate and more dependable estimates of state, and wholly new IMU layouts that seek to reduce error. New types include: gyro-free, timing, wireless, distributed redundant IMUs, and IMUs that incorporate MEMS components for miniaturization in general. This review paper highlights these new research directions and lays out the design and experimental implementation of a complementary filter for inertial measurement.

\end{abstract}

\section{INTRODUCTION}

Interest in inertial measurement began to accelerate in the fifties alongside the U.S. missile program. The LGM-30 Minuteman in particular is notable for driving advances in both inertial measurement and the miniaturization of computers: inertial navigation is difficult without computers. Initially, the process of fueling ballistic missiles for launch took around 30 minutes, which was about equal to the time needed to ready the mechanical inertial navigation system \cite{missile}. However, Minuteman was designed to be launched quickly, and the advent of solid fuel engines eliminated the fueling delays. This left the guidance system as the limiting factor. To keep launch times down, the guidance system would have to be ready at all times: this was a problem because of its mechanical components, Electronic computers and guidance systems were proposed as a solution, but transistor-based computers were not sufficiently reliable at the time. Interestingly, the Air Force’s successful, multimillion dollar effort to improve transistor production directly translated to a substantial reduction in miniature computer costs and heavily impacted the fledgling electronics industry \cite{missile}.

Inertial Measurement Units (IMUs) are self-contained acceleration and orientation-sensing devices which are comprised of triads of accelerometers, gyroscopes, and sometimes magnetometers. These triads are offset to provide input data in all three dimensions. IMUs are used to maneuver anything that moves, such as robots, aircraft, spacecraft, missiles, etc. In addition to navigation, IMUs are used to measure human orientation for applications such as motion capture, sports technology, and virtual reality. When combined with a Global Positioning System (GPS), they make up the widely used Inertial Navigation System (INS) \cite{strapdown}, which combines the strengths of both systems: an INS can be used to navigate in places where GPS satellites are unreachable due to electronic interference or physical obstruction. IMUs also provide more precise measurements (at scales close to 10 m) than GPSs, but suffer from an accumulation of error which must be periodically corrected by GPS data. In contrast to GPSs, which utilize triangulation and rely on continuous communication with external sources, IMUs may be completely self contained. Both gyro-free and traditional inertial measurement units benefit from being periodically reset by outside data sources such as a GPS.

\begin{figure}[ht]
\centering
\includegraphics[scale=.4]{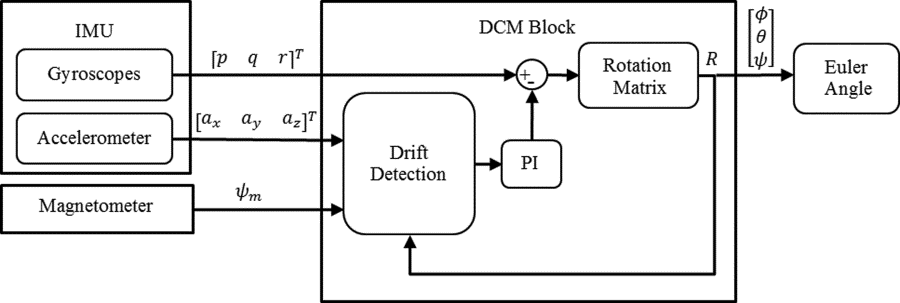}
\caption{Standard IMU configuration block diagram \cite{hybrid}}
\end{figure}

\begin{figure}[ht]
\centering
\includegraphics[scale=.2]{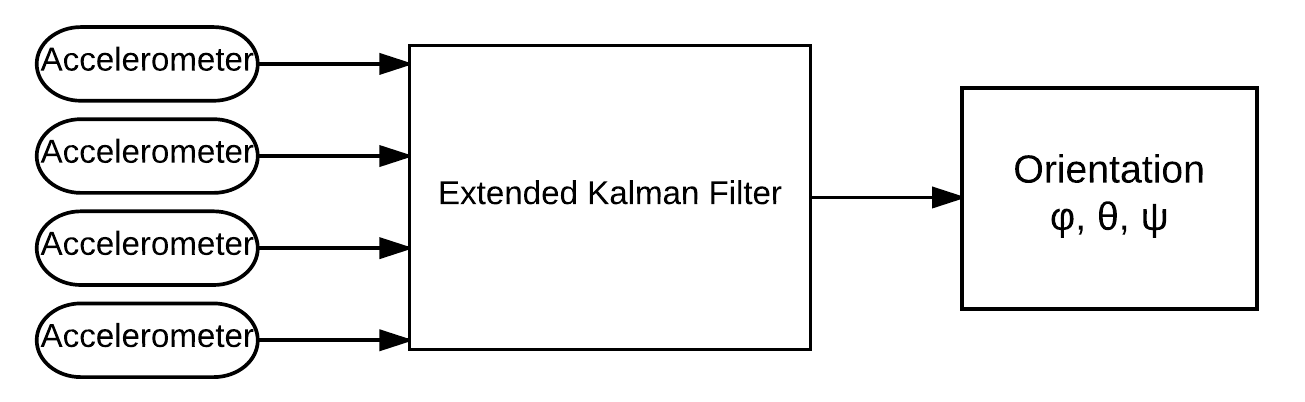}
\caption{Gyro-Free IMU Layout}
\end{figure}

\section{MATHEMATICAL BACKGROUND}

Estimation problems (such as inertial measurement) are fundamentally optimization problems in the sense that the goal is the minimization of the mean-square error. The current state-of-the-art for performing the task of orientation estimation relies upon Kalman filters, which take in a series of measurements over time and output estimates of unknown variables \cite{kalman}.  Kalman filters are widely used in diverse fields such as inertial navigation, signal processing, robotic motion planning, and modeling the human central nervous system. They are linear, discrete-time systems that estimate a state with the goal of minimizing the mean-square error. However, in many engineering applications, the system state dynamics are nonlinear and a KF may not be used: the solution to this problem is an Extended Kalman filter \cite{kalman}. The EKF is a form of KF that is derived from the linearization of the original system dynamics equations. A Kalman filter may be represented as follows:

\begin{figure}[ht]
\centering
\includegraphics[scale=.55]{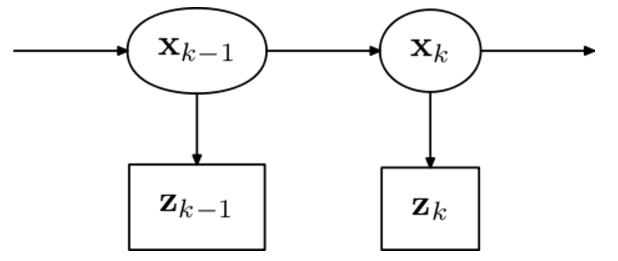}
\caption{Kalman Filter}
\end{figure}

\begin{equation}
{  {x}}_{{k}}=f({  {x}}_{{k-1}},{  {u}}_{{k-1}})+{  {w}}_{{k-1}}
\end{equation}
\begin{equation}
{  {z}}_{{k}}=h({  {x}}_{{k}})+{  {v}}_{{k}}
\end{equation}

\noindent Where $x_k$ is the state, $z_k$ is the observation, $w_{k-1}$ and $v_k$ are the process and observation noises and $u_{k-1}$ is the control vector.\\

One notable drawback to the Extended Kalman Filter is that if the initial estimate of the state is wrong, or if the process is modeled incorrectly, the filter may quickly diverge owing to its linearization. This is why accurate initialization of both the state and covariance matrices is essential. Other similar filtering methods that may be used are the Complementary Filter (used below), the Unscented Kalman Filter, the Kalman-Bucy Filter, and the particle filter.

\section{CURRENT PROBLEMS AND SOLUTIONS}

Major problems being tackled in the field of inertial navigation are error accumulation, size, weight, and power consumption. With respect to error, gyroscopes suffer from gradual drift, which increases proportionally to the square root of time. For this reason, they are often periodically corrected with a GPS. As outlined above, IMUs use gyroscopes combined with accelerometers. However, gyroscopes are theoretically not needed in order to output accurate orientation/attitude estimates \cite{gfvel}. Furthermore, gyroscopes contribute to a significant majority of the cost, weight, and power consumption of an IMU: it is for this reason that eliminating them would be desirable if accuracy could be maintained. On the other hand, it should be noted that MEMS gyroscope technology has also progressed, mitigating some of the aforementioned issues. Software and hardware solutions to the above problems exist, but are not yet optimal. For example, more accurate components may be used to reduce error, but this increases cost. Gyro-free IMUs reduce weight and cost but further development of their “software gyroscope” components is needed to demonstrate comparable or increased accuracy. The current state of the field is best summarized, “Although the concept of determining angular velocity out of the readings of multiple, displaced accelerometers is not new, apparently it has not yet matured into practical technologies.” \cite{gfins}. A central goal, then, in IMU technology is the development of a reliable method to estimate orientation from displaced accelerometers only.

Notable progress on this front came in the form of a 1994 inertial measurement method that uses six uni-axial accelerometers instead of nine, which was common before due to practical performance advantages \cite{gfstrapdown}. The configuration shown below uses the theoretical minimum of six uni-axial accelerometers \cite{isotropic} and has the advantage of providing favorable condition numbers for the state and estimation matrices. A low condition number is desirable because it provides a low upper bound for the error that may be introduced through estimation.

\begin{figure}[ht]
\centering
\includegraphics[scale=.4]{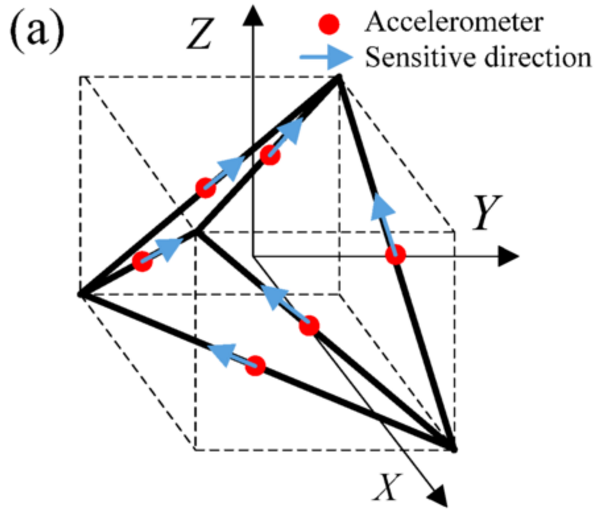}
\caption{Accelerometer Configuration \cite{maglev}}
\end{figure}

As of 1994, solely using accelerometers was found to greatly increases the error in orientation estimation when compared to using a gyroscope. However, it was noted then that accelerometers are suitable for short mission durations with high dynamic acceleration, have lower power requirements, and do not wear out like mechanical gyroscopes.\cite{gfstrapdown}

There has been recent progress in the field of gyro-free inertial navigation systems (GF-INSs) \cite{gfins}. First proposed in the sixties \cite{gfvel}, the GF-INS has the advantage of replacing costly, heavy, and power-intensive gyroscopes with accelerometers, which are undergoing rapid miniaturization as Micro Electromechanical Systems (MEMS) \cite{mems}. It remains to be seen if GF-INSs can be made as accurate as traditional systems.
A notable recent demonstration of working GF-INS is given by Cucci et. al. (2016). In their paper, they lay out a modified EKF called the $\omega$-filter which is used to estimate orientation from displaced accelerometer readings. Their system layout and results are shown below:

\begin{figure}[ht]
\centering
\includegraphics[scale=.8]{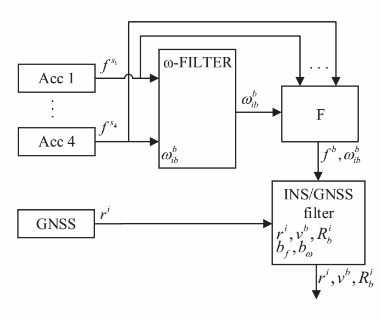}
\caption{Gyro-Free INS Layout \cite{gfins}}
\end{figure}

In the block diagram, the top half (outputting f and $\omega$) is very similar to the GF-IMU layout shown in the introduction. The plots at the end of this section show the error vs. time of both the GF-algorithm and a standard IMU. Evidently, the GF-algorithm is currently much noisier than the standard IMU - there is significant room for refinement in the algorithm.

\begin{figure}[!ht]
\centering
\includegraphics[scale=.9]{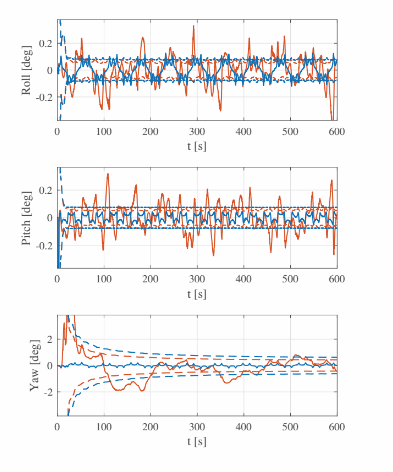}
\caption{Gyro-Free INS Experimental Results: Orientation Error vs. Time. Blue curves represent IMU results, while red curves represent the GF-algorithm \cite{gfins}.}
\end{figure}

Finally, DARPA is also currently developing Timing and IMU (TIMU) sensors that use an integrated timer as an added data source and are much smaller than current IMUs. "The TIMU goal is to develop a tactical-grade IMU, including simultaneous co-fabrication of 3 gyroscopes, 3 accelerometers and a resonator, in unprecedented cost, size, weight, and power." \cite{timu}. This design will be suitable for applications such as missiles, UAVs, UUVs, and other small vehicles that require long periods of autonomous operation.

\section{EXPERIMENT}

A complementary filter was chosen for the experiment due to its simplicity. The basic principle is to use the accelerometer data to compensate for the gyroscope drift (the gyro drifts over time but the accelerometer does not). Both signals are combined at different weights depending on how quickly the body is accelerating - at low dynamic accelerations, mainly accelerometer data is used, and vice versa. This is because accelerometer data is very good for determining orientation when the body is at rest (the gravity vector provides a reference), but this does not work well during acceleration so gyro data must be used. The equation for a complementary filter is as follows and is based off of the sensor fusion algorithm outlined by Madgwick\cite{madgwick}: 

\begin{equation}
{  {pitch}_{t}= pitchAccelerometer_{t} * (1-\gamma_{t}) + pitchGyroscope_{t} * \gamma_{t}}
\end{equation}

\noindent Where $\gamma$ varies with time and depends on the dynamic acceleration of the body.\\

The equation for $\gamma$ used in this filter was chosen to make $\gamma$ vary with acceleration - when acceleration is low, the pitch is determined mainly from the accelerometer data. The equation is as follows:
\begin{equation}
{{\gamma_{t} = (9.81 - acceleration_{z})_{normalized}}}
\end{equation}

The basic methods used to determine pitch from gyroscope and accelerometer data are as follows: gyroscopes return rotational velocity ($\omega$), which is integrated over time to determine pitch. Accelerometers return force readings along the x', y', and z' vectors (x', y', and z' being defined in reference to the rotating body) - while the body is at rest, it is possible to take the difference between the gravitational constant (g) (which equals the z' accelerometer reading when the body is level) and the actual z' reading. Then, pitch may be determined through geometry: 

\begin{equation}
{{pitchAcc = atan(a_{x}/\sqrt(a_{y} * a_{y} + a_{z} * a_{z}))}}
\end{equation}

\begin{figure}[!ht]
\centering
\includegraphics[scale=2.3]{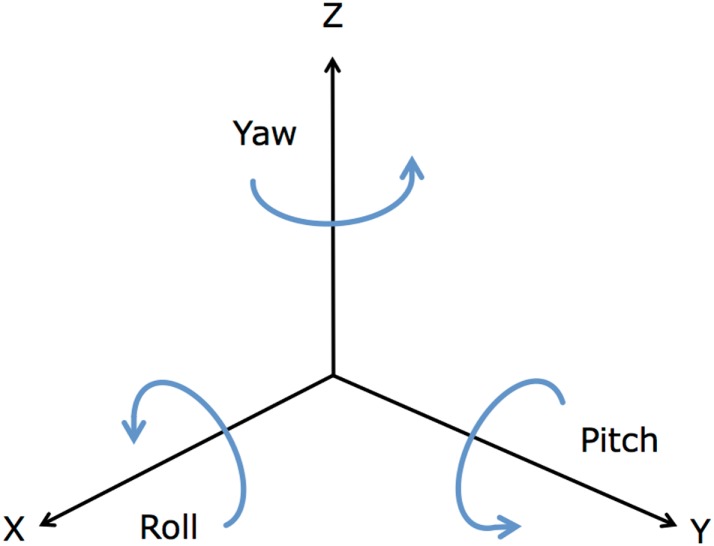}
\caption{Roll, Pitch, and Yaw}
\end{figure}

The frame of reference is defined in terms of roll, pitch and yaw. 
\pagebreak

\begin{figure}[!ht]
\centering
\includegraphics[scale=.6]{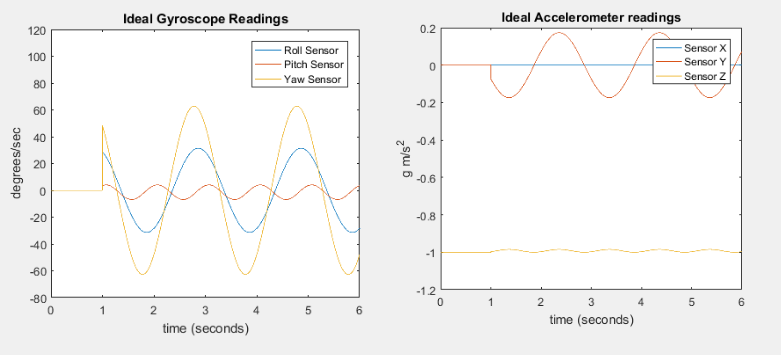}
\caption{Simulated IMU Data}
\end{figure}

\begin{figure}[!ht]
\raggedleft
\includegraphics[scale=.6]{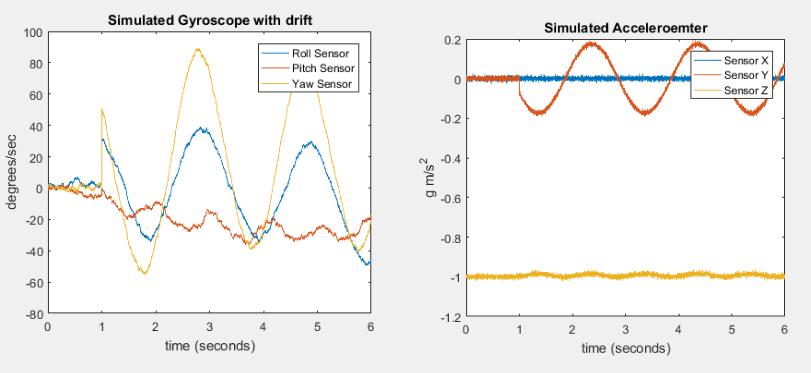}
\caption{Simulated Data with Noise}
\end{figure}

Figure 8 shows simulated IMU data from a virtual fish. The plots shown describe the "actual" path of fish, but the data being fed to the complementary filter has noise introduced to the accelerometers and gyroscopes. A random walk is also applied to the gyroscope. The simulated data with noise is shown in Figure 9 - the gyroscope experiences significant drift. 
\pagebreak

\begin{figure}[!ht]
\raggedleft
\includegraphics[scale=.6]{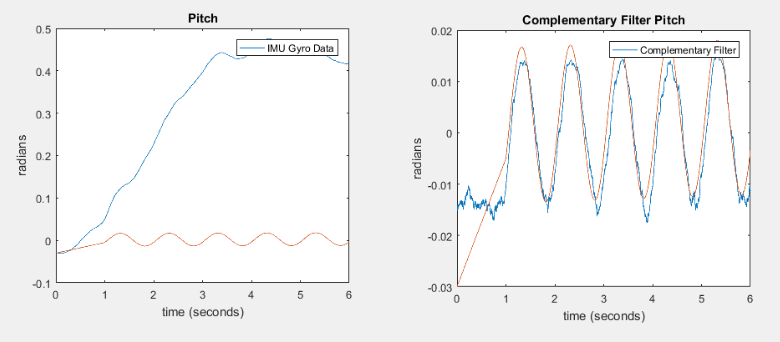}
\caption{Complementary Filter Results}
\end{figure}

Figure 10 shows the results from the complementary filter compared to the gyroscope data alone. Over time, the gyro drifts and becomes off by 0.3 radians over the course of 5 seconds. On the other hand, the complementary filter achieves much better results, closely tracking the true pitch. The complementary filter is behaving appropriately, smoothing the accelerometer data and compensating for the gyroscope drift.\\

Code can be found here: https://github.com/wtconlin/InertialSensors

\section{CONCLUSIONS}

Promising IMU research directions are gyro-free, timing, wireless, distributed redundant IMUs, and IMUs that incorporate MEMS components for miniaturization in general. IMU algorithms also show significant room for improvement - Kalman, EKF, Particle, and Complementary Filters are some of the most popular. The complementary filter for inertial measurement implemented here showed a significant increase in accuracy over gyroscope or accelerometer data alone and was very simple to implement as well as computationally inexpensive. Further improvements in the complementary filter algorithm outlined above will probably come from changes to the $\gamma$ function.

\bibliographystyle{plain}

\end{document}